\documentclass[conference]{IEEEtran}

\usepackage[dvips]{graphicx}
\usepackage{amsmath,amssymb}
\usepackage{algorithm}
\usepackage{algorithmic}
\usepackage{flushend}

\usepackage{url}

\usepackage[utf8]{inputenc} % Required for inputting international characters
\usepackage[T1]{fontenc} % Output font encoding for international characters

\usepackage{csquotes}
\newcommand{\dq}[1]{\enquote{#1}}

\begin{document}

\title{Keyphrase Generation: A Multi-Aspect Survey}

\date{}%date stay empty

\author{
\IEEEauthorblockN{Erion \c{C}ano, Ond\v{r}ej Bojar} 
\IEEEauthorblockA{Charles University \\ Prague, Czech Republic \\ \{cano, bojar\}@ufal.mff.cuni.cz\\}
%\and
%\IEEEauthorblockN{}
%\IEEEauthorblockA{\\ \\ \\}
}
\maketitle

\begin{abstract}
Extractive keyphrase generation research has been around since the nineties, but the more advanced abstractive approach based on the encoder-decoder framework and sequence-to-sequence learning has been explored only recently. In fact, more than a dozen of abstractive methods have been proposed in the last three years, producing meaningful keyphrases and achieving state-of-the-art scores. In this survey, we examine various aspects of the extractive keyphrase generation methods and focus mostly on the more recent abstractive methods that are based on neural networks. We pay particular attention to the mechanisms that have driven the perfection of the later. A huge collection of scientific article metadata and the corresponding keyphrases is created and released for the research community. We also present various keyphrase generation and text summarization research patterns and trends of the last two decades.   
\end{abstract}

%% keywords: Keyphrase Extraction, Abstractive Keyphrase Generation, Encoder-Decoder Framework, Novel Keyphrase Dataset, Literature Survey 

\section{Introduction}
\label{sec:intro}
A keyphrase or a keyword (here we use them interchangeably) is a short set of one or a few words that represent a concept or a topic covered in a document. They are commonly used to annotate articles or other documents and are essential for the categorization and fast retrieval of such items in digital libraries. A keyphrase string, on the other hand, is a set of comma-separated (other separators may be used as well) keyphrases associated with an article or a different type of object, describing the content and topical aspects of it. 
Because of their high importance and the need to process huge amounts of documents with missing keyphrases, KG (Keyphrase Generation) attracted high academic interest since the 90s. Some basic works of that time such as \cite{Witten1999}, \cite{Turney1999}, and \cite{Turney2000} used text features and supervised learning algorithms (popular at that time) to extract keywords from documents. Improved supervised methods like \cite{Wang2006}, \cite{Medelyan2009}, and \cite{Nguyen2010}, graph-based methods like \cite{Wan2008}, \cite{Rose2010}, and \cite{Bougouin2013}, or other unsupervised KG methods such as \cite{Nart2014} and \cite{Campos2018} were proposed in the 2000s. 
Extractive KG became so popular in the 2000s and early 2010s, that the entire research field was commonly called KE (Keyphrase Extraction). This success was mainly due to the simplicity and speed of the proposed solutions. There is still a serious flaw in extractive KG: the inability to produce absent keyphrases (predicted keyphrases that do not appear in the source text). Analyzing the most popular datasets, \cite{Meng2017} showed that present (predicted keyphrases that also appear in the source text) and absent keyphrases assigned by paper authors are almost equally frequent. Ignoring the later is thus a serious handicap.    
Motivated by the advances in sequence-to-sequence applications of neural networks, several studies like  \cite{Meng2017} or \cite{Zhang2018} started to explore AKG (Abstractive Keyphrase Generation). The encoder-decoder (or sequence-to-sequence learning) paradigm that was first utilized in the context of machine translation (e.g., in \cite{Cho2014}, \cite{Sutskever2014} or \cite{Bahdanau2014}) got quick adaption in related tasks such as text summarization (like in \cite{Chopra2016} and \cite{See2017}) or AKG. Since that time, AKG research took over and is today a vibrant field of study.
In this survey, we start by reviewing the most popular KE methods, specifically the supervised, the graph-based and the other unsupervised ones. We go on describing the popular existing keyphrase datasets and present OAGKX, a novel and huge collection of about 23 million metadata samples (titles, abstracts, and keyphrase strings) from scientific articles that is released online (\url{http://hdl.handle.net/11234/1-3062}). It can be used as a data source to train deep supervised KG methods or to create byproducts (other keyphrase datasets) from more specific scientific disciplines. 
Unlike similar recent reviews such as \cite{Papagiannopoulou2019}, \cite{Hasan2014}, or \cite{Siddiqi2015} that focus entirely on extractive KG (or KE), the main interest of this work in the more recent and technically advanced AKG studies which are examined in details. Particular attention is paid to the network structures and the enhancement mechanisms, as well as to the evaluation process the authors follow. We also describe certain research patterns that we observed such as the interesting analogy with similar developments in text summarization research. 

\section{Extractive Keyphrase Generation}
\label{sec:extractive}
Extractive keyphrase generation methods are simpler and appeared in the literature in the late 90s. They usually follow two steps. First, candidate phrases are selected from the document. Different strategies are latter applied to decide if each candidate is a keyphrase or not. The following subsections briefly describe the most popular extractive methods. More comprehensive and detailed reviews entirely focused on KE can be found in other surveys such as \cite{Papagiannopoulou2019}.  
\subsection{Supervised Methods}
\label{ssec:supervised}
One of the first studies that considered KG as a supervised learning problem was \cite{Turney2000}. In that study, the author experimented with texts from journal articles, email messages, and Web pages. Some of the used features were \emph{word frequency}, \emph{phrase length}, \emph{number of words in phrase}, etc. C4.5 decision tree algorithm of \cite{Quinlan1993} was utilized as a classifier, in combination with a bagging procedure based on random sampling with replacement presented at \cite{Quinlan1996}. The author also experimented with GenEx, an algorithm described in \cite{Turney1999} that was specifically designed to extract document keyphrases. He concluded that domain knowledge is highly valuable in the keyphrase extraction process and GenEx (using that knowledge) performs significantly better than C4.5 (not using it). This work encouraged other researchers to develop supervised learning methods for solving KE problems.    
Almost at the same time, KEA (Keyphrase Extraction Algorithm), a language-independent supervised KG algorithm was developed and presented in \cite{Witten1999}. It uses features like \emph{TF-IDF} and \emph{first occurrence} and then applies Na\"ive Bayes classifier to determine if candidate phrases are keyphrases or not. Authors evaluate KEA using NZDL dataset (see Table~\ref{table:publicdataStats}) and report that it is able to correctly identify one or two of the first five author keyphrases.
The development of Maui, a similar algorithm presented in \cite{Medelyan2009} was a further step forward. Maui extends KEA in several ways. It combines more feature types and exploits Wikipedia articles as a source of linguistic knowledge. Furthermore, Maui can work well with both Na\"ive Bayes and bagged decision tree classifiers. 

Some attempts explore various feature setups for improving the existing methods. The author in \cite{Hulth2003} investigates the role of additional features like n-grams, \emph{noun phrases}, \emph{POS tags}, etc. She concludes that using words or n-grams that match \emph{POS tag} patterns increases the recall compared to the usage of n-grams only. Furthermore, according to her results, the syntactic information of the POS tags is also important for optimizing the number of keyphrases assigned to each document. The author also creates and uses Inspec, a dataset of scientific paper metadata. Other studies like \cite{Wu2007} or \cite{Bhowmik2008} that followed also experimented with scientific paper texts, a practice that is common even today. 

The logical structure of a scientific article is defined in \cite{Mao2003} as the hierarchy of its logical components like title, list of authors, abstract and sections. Authors of \cite{Nguyen2010} use that logical structure to build WINGNUS by limiting the number of identified candidate phrases. They further use different features like \emph{length of phrases}, \emph{typeface} and \emph{position} (in title, introduction, etc.) for training the Weka implementation of Na\"ive Bayes (presented in \cite{Hall2009}) to select the best candidates. Authors conclude that using the logical structure of the scientific articles yields superior performance over methods that do not consider that information.

In \cite{Krapivin2010} they experimented by adding syntactic relations extracted with the dependency parser of \cite{Nivre2007}. They also tried different classifiers like Support Vector Machines of \cite{Cortes1995} and Random Forests of \cite{Breiman2001}. According to their results, the NLP-based features improve F$_1$ scores of all the tested methods. They also concluded that Random Forest is a good trade-off between keyphrase quality and generation speed.     

There were also a few studies that applied neural network structures to perform extractive KG. In \cite{Wang2006}, for example, they used a feed-forward neural network as a classifier and paid particular attention to title headings (also subheadings) and phrase repetitions. Authors of \cite{Villmow2018}, on the other hand, utilized a more complex neural network structure based on LSTMs (Long Short-Term Memory) to build an end-to-end keyphrase extraction system that eliminates the need for manual engineering of statistical features. 

\subsection{Graph-Based Methods}
\label{ssec:graph}
From the unsupervised extractive KG methods, those based on graph computations are the most numerous. In \cite{Mihalcea2004} they introduced TextRank, a graph-based ranking method inspired by the PageRank algorithm of \cite{Brin1998}. They implement the idea of \dq{voting}: a vertex that represents a word or phrase (lexical unit) links to another one, casting a vote in the later. A higher number of votes to certain words or phrases suggests that they are more important. All lexical units of the source text are ranked this way. The returned keyphrases are constructed from the top N words.    
Authors of \cite{Wan2008} use the concept of the neighborhood of a given document: a set of similar documents that expands that document. They later employ PageRank on the local graph (of a single document) or the expanded graph (of the neighborhood) to rank the words and phrases. SingleRank and ExpandRank are the names of the corresponding methods they derive. The authors report that ExpandRank is significantly better than SingleRank for any size of the neighborhood.   
In \cite{Gollapalli2014} they follow a similar approach to formulate CiteTextRank. Authors use the documents citing the given document (citation network) to expand it and then they apply PageRank.
TopicRank defined in \cite{Bougouin2013} is another improvement over TextRank. It first clusters lexical units of the document according to their topic. Afterwards, it uses a graph-based ranking model to assign scores to the topic clusters. Finally, keyphrases are generated by picking one of them from each ranked cluster. 
One of the fastest available KE methods is RAKE proposed in \cite{Rose2010}. Authors first remove punctuation and stop words and then create a graph of word co-occurrences. Candidate words are scored based on the degree and frequency of each word vertice in the graph. The top-scoring ones are returned as keyphrases. Authors report that RAKE achieves higher precision and similar recall when compared with other graph-based methods like TextRank.   
PositionRank is yet another graph-based KE approach recently proposed by \cite{Florescu2017}. They construct a word-level graph where they incorporate information from positions of all word occurrences. PageRank is later used to score the words and phrases. Authors show that using positions of all word occurrences works better than using the first occurrence of each word only. 

\subsection{Other Methods}
\label{ssec:other}
Besides the two categories above, there are also other unsupervised methods that are not graph-based. They mostly utilize clustering and various similarity measures to find the best keyphrases.  
A very simple scheme uses TF-IDF to compute scores and rank text phrases of the entire document. This raw approach is one of the most frequent baselines in other studies that propose KG methods.

Authors of \cite{HaCohen-Kerner2003} proposes another basic approach based on term frequencies and stopword filtering.  
In \cite{Nart2014} they argue that KG systems should be unsupervised and domain-independent. They build a KG system based on loosely structured ontologies. 
Authors of \cite{Jo2015} rely on Deep Belief Networks described in \cite{Hinton2006} to capture the intrinsic representations of documents and using them to extract keyphrases. 

Another peculiar approach is the one by \cite{Liu2011} who consider keyphrasing as a form of translation from the language of the document to the language of keyphrases. They use word alignment from statistical machine translation to learn matching probabilities between document words and keyphrase words.

Statistical language models are also used by \cite{Tomokiyo2003} who utilize Kullback-Leibler divergence described in \cite{Vidyasagar2010} to create a single score (including phraseness and informativeness) for ranking extracted phrases.  
YAKE! presented in \cite{Campos2018} is another example of an unsupervised and feature-based extractive KG solution. They utilize features like \emph{casing}, \emph{word position}, \emph{word frequency}, and more, combined in a complex scoring function that is used to yield the ranked keyphrases.

There is also a recent attempt in \cite{key2vec} to use the concept of word embeddings in the context of the unsupervised KE. Authors propose Key2Vec, a method for training phrase (multi-word) embeddings which are used to represent the candidate keyphrases and build the thematic representation of the document. The candidate keyphrases are later ranked based on their thematic relation with the document using the theme-weighted PageRank algorithm of \cite{PageRank2}. 

The many extractive (supervised, graph-based or other) KG methods described in this section are complementary and may be used in different scenarios and for different purposes. To ease their implementation and benchmarking, the author of \cite{Boudin2016} created PKE, a Python toolkit available online (\url{https://github.com/boudinfl/pke}). It implements many of the above methods, offers pretrained and ready to use KE models and can also be easily extended to implement or benchmark new methods. 

\section{Keyphrase Datasets}
\label{sec:datasets}

\subsection{Popular Corpora}
\label{ssec:scidata}
The recent open data initiatives and data science competitions have encouraged the creation and sharing of more and more datasets. There are papers like \cite{Movies2015} that release data about movies, \cite{Songs2017} about music, \cite{Books2017} about books and \cite{Cano2015} that describes data of other object categories. The computational linguistics or natural language processing datasets consist of various text collections that are used to solve particular tasks. In the realm of KG, the most popular in the literature are the collections of scientific articles shown in Table~\ref{table:publicdataStats}.

Inspec is one of the earliest datasets, released in \cite{Hulth2003} where the role of various linguistic features in KE is explored. It consists of 2000 paper titles (1500 for training and 500 for testing), abstracts and keywords from journals of Information Technology, published from 1998 to 2002.

One of the smallest is NUS of \cite{Nguyen2007}, consisting of 211 conference papers. Each paper has two sets of keyphrases: one set by the authors and a second that was created by volunteer students. 
Another small dataset is SemEval (or SemEval-2010) described in \cite{Kim2010}. It is composed of 244 papers, 144 for training and 100 for testing. They were collected from ACM Digital Library and belong to conference and workshop proceedings.  

Krapivin, the dataset released in \cite{Krapivin2010} has the advantage of providing full paper texts together with the corresponding metadata. There is a total of 2304 Computer Science articles published by ACM from 2003 to 2005. The parts of each text such as title, abstract and sections are separated and marked to ease the extraction of various keyphrases. 

The most popular KG dataset of the recent years is probably KP20k released in \cite{Meng2017}. It consists of 567830 Computer Science articles, 527830 for training, 20K for validation and 20K for testing. KP20k has been used for training and evaluating various recent abstractive methods.  
The biggest keyphrase dataset is probably OAGK
recently released in \cite{Cano2019}. It contains 2.2M titles, abstracts and keyphrase strings of scientific papers from different disciplines. 

The above scientific paper datasets are summarized in Table~\ref{table:publicdataStats}. There are also a few more datasets of other document types, but they are less popular in the literature. One of them is NZDL, a collection of 1800 Computer Science technical reports, 1300 for training and 500 for testing. It is described in \cite{Witten1999}. Authors use it to benchmark KEA, their extractive method which was one of the first.

From the news domain, the DUC (or DUC-2001) dataset of \cite{Wan2008} is somehow popular. It consists of 308 news articles and 2048 keyphrase labels and has been used in a few extractive and abstractive KG methods. 
In \cite{Zhang2016} they create a dataset of about 147K tweets and their corresponding tags. Authors use it to evaluate their model for hashtag prediction.  
Authors of \cite{Yih2006} use a dataset of 815 Web pages and the corresponding extracted keywords for addressing advertisements.   

The two most recent datasets are probably StackExchange (post topics) and TextWorld (game observations and commands) created and used by \cite{Yuan2018}. Similar datasets can be found in other works like \cite{Dredze2008}, \cite{Grineva2009} or \cite{Hammouda2005}.  
\begin{table}[!t]
\caption{Public keyphrase datasets}
\centering
% \normalsize
\small
\begin{tabular}{|l|l|l|c|}
\hline
\bf Reference & \bf Name & \bf Content & \bf \# Docs  \\ [0.1ex] 
\hline
\cite{Hulth2003} Hulth & Inspec & Papers & 2000 \\ [0.07ex]
% \hline
\cite{Nguyen2007} Nguyen & NUS & Papers & 211 \\ [0.07ex]
\cite{Kim2010} Kim & SemEval & Papers & 244 \\ [0.07ex]
~\cite{Wan2008} Wan & DUC & News & 308 \\ [0.07ex]
\cite{Krapivin2010} Krapivin & Krapivin & Papers & 2304 \\ [0.07ex]
\cite{Zhang2016} Zhang & Twitter & Tweets & 147K \\ [0.07ex]
\cite{Meng2017} Meng & KP20k & Papers & 567K\\ [0.07ex]
~\cite{Witten1999} Witten & NZDL & Reports & 1800 \\ [0.07ex]
\hline
\end{tabular}
\label{table:publicdataStats}
\end{table}

\subsection{A Novel and Huge Data Collection}
\label{ssec:oagkx}
Experimenting with keyphrases of scientific papers seems an ongoing trend that is greatly motivated by the availability of data in online academic repositories. Following the examples of \cite{Cano2019} and \cite{Cano2019b}, we took the initiative to produce an even larger collection of scientific paper keywords, titles and abstracts. Exploiting the whole data of Open Academic Graph (described in \cite{Tang2008} and \cite{Sinha2015}), we retrieved \emph{keywords}, \emph{title} and \emph{abstract} data wherever they were available. A language filter was applied to remove every text record not in English. We also lowercased and utilized Stanford CoreNLP of \cite{Manning2014} to tokenize the \emph{title} and \emph{abstract} texts.   
Since there were several articles with very short or very long text fields (outliers), we removed any record with a title not within 3-25 tokens, abstract not within 50-400 tokens or keyphrase strings not within 2-60 tokens. We also removed records with a number of keyphrases now within 2-12. The obtained dataset is OAGKX, a collection of about 23 million article metadata records. 

Some basic statistics regarding the distribution of tokens in \emph{title}, \emph{Abstract} and \emph{Keywords} fields of the articles can be found in Table~\ref{tab:tokenstats}. As we can see, the average lengths are about 13 tokens for the titles, 175 tokens for the abstracts, and 12 tokens for the keyphrase strings (standard deviation is given in parenthesis). We also computed the token overlaps between abstracts and titles, and between abstracts and keyphrase strings. The overlap $o(s, t) = \frac{|\{s\} \cap \{t\}|}{|\{t\}|}$ between two token vectors $s$ (source) and $t$ (target) is the fraction of unique tokens in $t$ that overlap with a source token in $s$. As we can see, there is high repetition of abstract words, both in titles (78\,\%) and in keyphrases (68\,\%).

\begin{table}[!t]
\caption{Token statistics of OAGKX}
\small 
\centering
\begin{tabular}{|l|l|}
       \hline
       \bf Attribute & \bf ~~Title \qquad~ Abstract ~~~~~ Keywords \\
       \hline
       Total & ~ 290\,M \qquad~~~\, 4\,B \qquad\qquad~ 270\,M \\
      \hline
%       Unique & \quad 3.2\,M \qquad\, 50/400 \qquad~~ 2/60 \\
%       \hline
       Min\,/\,Max & \quad 3\,/\,25 \qquad~\, 50\,/\,400 \qquad\quad~ 2\,/\,60 \\
      \hline
       Mean & 12.8\,(4.9) ~~ 175.1\,(86.5) \quad~~ 11.9\,(7.5) \\
		\hline
%       Jindex & \quad~ 7.1\,\%\,(4\,\%) \quad~ 6\,\%\,(4.8\,\%) \\
% 		\hline
       Overlaps & ~~~\, 78\,\%\,(17\,\%) \quad~~~ 68\,\%\,(25\,\%) \\
% 		\hline
%       Missing & \qquad\quad -- \qquad\quad~~~ 47\,\%\,(28\,\%) \\
%		\hline
%       Total size & 22\,674\,436 title-abstract-keywords \\ 
% 		\hline
% 	   Used (tr/v/ts) & \quad~ 2\,000\,000 / 10\,000 / 10\,000 \\
      \hline
\end{tabular}
\label{tab:tokenstats}
\end{table}

We further observed the distribution of keyphrases. The corresponding statistics are shown in Table~\ref{tab:keystats}. There is a total of about 133 million keyphrases with an average of about 6 in each article. The minimal and maximal of keyphrases in each record is 2 and 12 respectively. In KG experiments, it is also important to check the frequencies of the keyphrases that are present and absent in the source texts. The present fraction $p(s, k) = \frac{|k~\cap~s|}{|k|}$ is the fraction of the  keyphrases $k$ that do appear in the source text $s$. The absent fraction $a(s, k) = \frac{|k| - |k~\cap~s|}{|k|}$ is the its complement, or in other words the fraction of the keyphrases $k$ that do not appear in the source text $s$. As we can see, OAGKX present and absent keyphrases are almost equally frequent (52.7\,\% vs. 47.3\,\%). This is in line with the observation of \cite{Meng2017}.

Using three extractive methods described in Section~\ref{sec:extractive}, We performed some preliminary experiments with OAGKX data. We picked YAKE!, RAKE and TopicRank which are simple and used them with their default parameters in each implementation. Given that they are unsupervised and require test data only, we picked a big test cut of 100K samples from the entire OAGKX. In addition to the preprocessing steps described above which were performed on entire OAGKX collection, we also replaced digit symbols with \# and joined each title and abstract in common source string. The length of this source string was limited to 260 tokens (a paper abstract and the title should not be longer). 

For the evaluation, we used F$_1$ score of full matches between predicted keyphrases from each method and those available in the data record (author keyphrases). We computed F$_1$ scores on top 5, top 7 and top 10 returned keywords. Before comparing, both sets of terms were stemmed with Porter Stemmer and duplicates were removed. The obtained results are presented in Table~\ref{table:extractiveOAGKX}. As we can see, the best of the three methods is YAKE, with top F$_1@10$ score of 21.86\,\%. We also observed that RAKE was considerably faster than the two other methods.   
To have an idea about the topic distribution of OAGKX articles, we inspected a few randomly picked data records. We noticed that they belong to various scientific disciplines, with medicine (and its research directions) being dominant. There are also many papers about economics, social sciences or different technical disciplines. To our best knowledge, this is the biggest available collection of scientific paper data and the corresponding keyphrases. The value of OAGKX is thus twofold: (i) It can supplement the existing datasets if more training data are required. (ii) It can serve as a data source for creating scientific article subsets of more specific scientific disciplines or domains.       
% 

%% Table with dataset statistics
% \begin{table}[!t]
% \caption{OAGKX length statistics in \# tokens}
% \centering
% % \normalsize
% \small
% \begin{tabular}{|l|c|c|c|}
% \hline
% \bf Attribute & \bf Title & \bf Abstract & \bf Keyword  \\ [0.1ex] 
% \hline
% Total tokens & 344M & 5.5B & 248M \\ [0.07ex]
% % \hline
% Min. Length & 1 & 1 & 1 \\ [0.07ex]
% Max. Length & 815 & 304K & 28K \\ [0.07ex]
% Av. Length & 13.2 & 203.9 & 8.5 \\ [0.07ex]
% \hline
% \end{tabular}
% \label{table:dataStats}
% \end{table}
% %

%% 
\begin{table}[!t]
\caption{Keyword statistics of OAGKX}
\small
\centering
\begin{tabular}{|l|c|}
       \hline
      \bf Attribute & \bf Value \\
      \hline
       Total & 133\,295\,056 \\
    %   \hline
       Min\,/\,Max & 2\,/\,12 \\
% 		\hline
       Mean & 5.9\,(3.1) \\
% 		\hline
%       Jindex & 6.7\,\%\,(3.9\,\%) \\
%		\hline
       Present & 52.7\,\%\,(28.3\,\%) \\
% 		\hline
       Absent & 47.3\,\%\,(28.3\,\%) \\
       \hline
\end{tabular}
\label{tab:keystats}
\end{table}

\section{Abstractive Keyphrase Generation}
\label{sec:seq2seq}
In this section, the recent AKG methods based on the encoder-decoder framework are examined in detail. Table~\ref{table:abstractiveTable} summarizes some of their neural network properties, together with the evaluation data and metrics used by the authors. 
\subsection{Basic Neural Network Models}
\label{ssec:basic}
The authors of \cite{Zhang2016} were among the first to try RNNs (Recurrent Network Networks) for generating keyphrases (actually hashtags) of tweets. They adopt a joint-layer RNN with two hidden layers and two output ones. The latter are combined to form the objective layer (keyword or not). Authors build and refine a big dataset of tweets and the corresponding hashtags (keywords in this context) for evaluating their method. The basic LSTM of \cite{Hochreiter1997} and AKET, a tool for keyword extraction on tweets described in \cite{Marujo2015} are used as comparison baselines. Superior scores of 80.74\,\%, 81.19\,\% and 80.97\,\% are reported in terms of P (Precision), R (Recall) and F$_1$ respectively.
Another important work is \cite{Meng2017}, the first to adapt the encoder-decoder framework for AKG. Their CopyRNN model has an encoder that creates a hidden representation of the source text and a decoder that generates the keyphrases based on that representation. They employ a bidirectional GRU of \cite{Cho2014} as the encoder and a forward GRU as the decoder. Keyphrase generation involves a beam search described in \cite{Dahlmeier2012} with max depth 6 and beam size 200. The attention mechanism of \cite{Dahlmeier2012} and copying mechanism of \cite{Gu2016} are implemented to improve performance and alleviate the out-of-vocabulary words problem.  

Authors evaluate CopyRNN on Inspec, Krapivin, NUS and SemEval and KP20k (IKNSK for short) datasets. Comparing with previous extractive approaches, they report state-of-the-art results in terms of F$_1@5$ (0.328 on KP20k) and F$_1@10$ (0.255 on KP20k) scores for present keyphrases. They also report top scores on R@10 and R@50 for absent keyphrases. Their work created a roadmap of using the encoder-decoder framework for AKG that has been followed by many other researchers in these last three years. 
In \cite{Zhang2017} they tried to optimize the speed of CopyRNN building CopyCNN made up of CNNs (Convolutional Neural Networks) which work in parallel. CNN layers are stacked on top of each other to process variable-length input text representations and gated linear units are used as the non-linearity function, same as in \cite{Dauphin2017}. They also use position embeddings combined with input word embeddings to preserve the sequence order. Authors test their method using IKNSK and compare against several extractive methods and CopyRNN. They report slightly higher performance scores (in F$_1@5$, F$_1@10$, R@10, and R@50) compared to CopyRNN of \cite{Meng2017}. Their model is also considerably faster, with generation times at least 6.2x lower. 
Furthermore, authors in \cite{Zhang2018} tried to improve another aspect of CopyRNN, handling of keyword repetitions during generation. They build their model (CovRNN) utilizing a bidirectional GRU for encoding and a forward GRU for decoding. To consider the correlation of the generated target keyphrases with each other (avoiding repetitions), they implement the coverage mechanism of \cite{Tu2016}. Same data (training on KP20k and evaluation on IKNSK) setups are used. The authors compare against extractive methods and CopyRNN. They report slightly better results compared to CopyRNN on both present (using F$_1@4$ and F$_1@8$) and absent (using R@10 and R@50) keyphrases.
\begin{table}[!t]
\caption{KE scores on OAGKX (100K)}
\centering
% \normalsize
\small
\begin{tabular}{|l| l l c|}
\hline
\bf Method & \bf F\textsubscript{1}@5 & \bf F\textsubscript{1}@7 & \bf F\textsubscript{1}@10 \\ [0.1ex] 
\hline
YAKE! & 19.27 & 21.49 & 21.86 \\ [0.07ex]
RAKE & 14.39 & 17.51 & 18.22 \\ [0.07ex]
TopicRank & 16.68 & 20.12 & 20.14 \\ [0.07ex]
\hline
\end{tabular}
\label{table:extractiveOAGKX}
\end{table}

\subsection{Enhanced and Hybrid Solutions}
\label{ssec:enhanced}
Many works followed, improving different aspects of AKG. Authors of \cite{Chen2018a} propose a solution for handling repetition and increasing keyphrase diversity. Besides using coverage, they also implement a review mechanism that considers the source context as well as a target context (collection of hidden states) before predicting (decoding) the next keyphrase. Same as above, they implement their model (CorrRNN) with bidirectional GRU, forward GRU and beam search. They utilize the training part of KP20k and evaluate on NUS, SemEval and Krapivin datasets, comparing against several extractive methods and CopyRNN. Given that keyphrase diversity is important, besides the typical F$_1$ and R metrics, they also utilize $\alpha$-NDCG of \cite{Clarke2008}. The authors report improvements on all reported metrics. Peak scores of 0.318 in F$_1@5$ and 0.278 in F$_1@10$ are reached on Krapivin dataset. They also assess the generalization ability of their model by training it with articles and testing it on news using DUC dataset.      
\begin{table*}[ht]
\caption{Summary of AKG model properties. IKNSK = \{Inspec, Krapivin, NUS, SemEval-2010, KP20k\}, NSK =  \{NUS, SemEval-2010, Krapivin\}, IKK = \{Inspec, Krapivin, KP20k\}, GT = Generation Time.} 
\small 
\centering      
\setlength\tabcolsep{7pt}  
\begin{tabular}
{| l | l c c c | l l |}
% \toprule
\hline
& \multicolumn{6}{|c|}{\textbf{Method} \qquad\qquad\qquad\qquad\qquad\qquad\qquad~ \textbf{Evaluation}} \\ [0.42ex] 
\textbf{\quad Reference} & \textbf{~~Network} & \textbf{Att} & \textbf{Copy} & \textbf{Cov} &
\textbf{Data} & \textbf{Metrics} \\ [0.12ex] 
% \midrule
\hline
\cite{Zhang2016} Zhang2016 & joint-layer, RNN & - & - & - & Tweets & Precision, Recall, F$_1$ \\ [0.07ex]
\cite{Meng2017} Meng2017 & Enc-Dec, GRU & \checkmark & \checkmark & - & IKNSK & F$_1@5$, F$_1@10$, R@10, R@50 \\ [0.07ex]
\cite{Zhang2017} Zhang2017 & Enc-Dec, CNN & \checkmark & \checkmark & - & IKNSK & F$_1@5$, F$_1@10$, R@10, R@50, GT \\ 
\cite{Zhang2018} Zhang2018 & Enc-Dec, GRU & \checkmark & \checkmark & \checkmark & IKNSK & F$_1@4$, F$_1@8$, R@10, R@50 \\ [0.07ex]
% \hline
\cite{Chen2018a} Chen2018a & Enc-Dec, GRU & \checkmark & \checkmark & \checkmark & NSK & F$_1@5$, F$_1@10$, R@10, N@5, N@10 \\ [0.07ex]
\cite{Ye2018} Ye2018 & Semisup, LSTM & \checkmark & \checkmark & - & IKNSK & F$_1@5$, F$_1@10$, R@10 \\ [0.07ex]
\cite{Chen2018b} Chen2018b & Enc-Dec, GRU & \checkmark & \checkmark & - & IKNSK & F$_1@5$, F$_1@10$, R@10, R@50 \\ [0.07ex]
\cite{Chen2019} Chen2019 & Hybrid, GRU & \checkmark & \checkmark & - & IKNSK & F$_1@5$, F$_1@10$, R@10 \\ [0.07ex]
\cite{Misawa2019} Misawa2019 & MultiDec, GRU & \checkmark & \checkmark & \checkmark & IKK & F$_1@5$, F$_1@10$, dist1, dist2 \\ [0.07ex]
\cite{Wang2019} Wang2019 & NTM, GRU & \checkmark & \checkmark & - & Blogs & F$_1@1$, F$_1@3$, F$_1@5$ \\ [0.07ex]
\cite{Yuan2018} Yuan2018 & catSeq, LSTM & \checkmark & \checkmark & - & IKNSK & F$_1@5$, F$_1@10$, F$_1@M$, F$_1@V$ \\ [0.07ex]
\cite{Chan2019} Chan2019 & RL, GRU & \checkmark & \checkmark & - & IKNSK & F$_1@5$, F$_1@M$ \\ [0.07ex]
% \bottomrule
\hline
\end{tabular} 
\label{table:abstractiveTable}
\end{table*}
All the above methods are supervised and depend on labeled training data which are not available for certain domains. In \cite{Ye2018} they try to overcome this limitation using two approaches. In the first one, they tag unlabeled documents with synthetic keyphrases obtained from unsupervised methods and use them for model pretraining. The pretrained model is later tuned on the labeled data. In the second one, they use multitask learning by combining the task of AKG based on labeled data with the task of title generation (a form of text summarization) on unlabeled data.  

Both tasks are implemented with a bilinear LSTM as the encoder and a plain LSTM as the decoder. In the multitask learning case, the encoder is shared by the two tasks wheres the decoders are different. Authors use KP20k as a source of labeled and unlabeled data and evaluate on IKNSK. A cross-domain test with news data (DUC dataset) is also performed. Their models outrun CopyRNN on all reported metrics (F$_1@5$, F$_1@10$ and R@10) reaching a peak score of 0.308 in F$_1@5$ on KP20k test set. 
Authors of \cite{Chen2019} try to inject the power of extraction and retrieval into the encoder-decoder framework. A neural sequence learning model is used to compute the probability of being a keyword for each word in the source text. Those values are later used to modify the copying probability distribution of the decoder, helping the later to detect the most important words. They also use a retriever to find documents annotated similarly which provide external knowledge for the decoder and guide the generation of the keyphrases for the given document. Finally, a merging module puts together the extracted, retrieved, and generated candidates, producing the final predictions. The authors use the same data and evaluation setup as above. They report superior scores of 0.317 in F$_1@5$ and 0.282 in F$_1@10$ for present keyphrases as well as significant improvements in R@10 scores for absent keyphrases.     
Furthermore, in \cite{Chen2018b}, they emphasize the important role of article title which indeed can be considered as a high-level summary of the text. Their solution (TG-Net) uses a complex encoder made up of three main parts. First, a bidirectional GRU is used to separately encode the source text (abstract + title) and the title in their corresponding contextual representations. Second, a matching layer catches the relevant title information for each context word using their semantic relation. Finally, another bidirectional GRU merges the original context and the gathered title information into a final title-guided representation. The decoder is similar to the ones described above, equipped with attention and copying. The authors train with KP20k and test on IKNSK. They report important gains over CopyRNN and CopyCNN on present keyphrases, with top scores 0.372 in F$_1@5$ and 0.315 in F$_1@10$ on KP20k test set. They also report significant improvements in absent keyphrases (higher R@10 and R@50 scores).     
An attempt to improve KG diversity is found in \cite{Misawa2019} where their method produces keyphrases one at a time, considering the formerly generated keyphrases. This is achieved by using multiple decoders (each of them generates only one keyphrase) that focus on different words of the source text by subtracting the attention value derived from the previous decoder. As a result, beam searches of beam size 1 are used to get the top keyphrase from each decoder and coverage is used to have diverse words in each keyphrase. The authors train their model with KP20k (the train split) and test on Inspec, Krapivin, and KP20k (the test split). They report improvements on keyphrase diversity measured using distinct-1 and distinct-2 metrics described in \cite{Li2016}.
In \cite{Wang2019} they create another hybrid system that infuses topical information into the encoder-decoder framework. They use an NTM (Neural Topic Model) for grasping the latent topic aspects of the input text. The later go into the decoder, together with the context representation of the input obtained by the encoder. Their learning objective is modified accordingly to balance the effects of the NTM and the KG encoder-decoder. Authors conduct experiments on blog data such as Twitter, Weibo (a Chinese microblogging website) and StackExchange. They compare tag prediction of their method against various previous methods such as CopyRNN, TG-Neg, and CorrRNN, reporting considerable improvements in terms of F$_1@1$, F$_1@3$ and F$_1@5$ scores. 
All the above works generate a fixed number of keyphrases per document. This is not optimal and realistic. In real scientific literature, different documents are paired with keyphrase sets of different lengths. To overcome this limitation and further improve the diversity of the produced keyphrases, authors of \cite{Yuan2018} propose a seq2seq generator equipped with advanced features. They first join a variable number of key terms as a single sequence and consider it as the target for sequence generation (sequence-to-concatenated-sequences or catSeq). By decoding a single of those sequences for each sample (e.g., taking top beam sequence from beam search) their model can produce variable-length keyphrase sequences for each input sample. 

For a higher diversity in output sequences, they apply orthogonal regularization on the decoder hidden states, encouraging them to be distinct from each other. Authors use the same data setup as in \cite{Meng2017} and compare against CopyRNN and TG-Net. Besides using F$_1@5$, F$_1@10$, they also propose two novel evaluation metrics: F$_1@M$, where $M$ is the number of all keyphrases generated by the model for each data point, and F$_1@V$, where $V$ is the number of predictions that gives the highest F$_1@V$ score in the validation set. Considerable improvements are achieved in terms of F$_1@10$ (top score 0.361), F$_1@M$ (top score 0.362) and F$_1@V$ (top score 0.362) on KP20k test set.
\subsection{Reinforcement Learning Perspective}
\label{ssec:rl}
Given that the above catSeq model tends to generate fewer keywords than the ground-truth, authors of \cite{Chan2019} reformulate it from the RL (Reinforcement Learning) perspective which has also been applied recently in several text summarization works like \cite{Chen2018fast}, \cite{Paulus2017} or \cite{Keneshloo2018} and similar seq2seq applications described in \cite{Keneshloo2018}. The model is stimulated to generate enough keyphrases employing an adaptive reward function that is based on recall (not penalized by incorrect predictions) in undergeneration scenarios and F$_1$ (penalized by more incorrect predictions) in overgeneration scenarios. They use GRU instead of LSTM but keep most of the other implementation details the same as those of \cite{Yuan2018}.  

The authors train on KP20k and test on IKNSK. They compare the RL-implemented catSeq, CopyRNN, and TG-Net against their original versions and report improvements from the RL implementation in all cases on both F$_1@5$ and F$_1@M$ with peak scores 0.321 and 0.386 respectively. The RL perspective is thus highly effective for enhancing existing AKG methods. Another contribution of their work is the novel comparison scheme they propose, with name variation sets for each ground-truth keyphrase. If a predicted keyphrase matches any name variation of a ground-truth keyphrase, it is considered as a correct prediction. 
\section{Keyphrasing Research Patterns}
\label{sec:patterns}
There are several patterns regarding technical and other aspects of research that show up from time to time. In this section, we briefly summarize some of such trends we identified in KG and TS (Text Summarization) research of the last two decades.  
\subsection{Experimental Patterns} % 
\label{ssec:commonSteps}
All of the primary studies we consulted perform some text preprocessing steps such as tokenization and lowercasing. Most papers do not report the tokenization utility they use. A few of them like \cite{Chen2018b} and \cite{Wang2019} report to have used Stanford CoreNLP of \cite{Manning2014} or NLTK (\url{www.nltk.org}) for tokenizing. It is also common to find KE studies like \cite{Boudin2016}, \cite{Hulth2003}, and \cite{Bougouin2013} that perform POS tagging and include the tags in the feature set they utilize.

A reduced vocabulary size is important to have decent AKG resutls within a reasonable computation time. For this reason, authors of many recent AKG studies like \cite{Meng2017}, \cite{Chen2018a}, \cite{Zhang2017}, \cite{Ye2018}, \cite{Chen2018b} and \cite{Chan2019} replace all digit tokens with the symbol $\langle$digit$\rangle$. 
Stemming is also commonly used in studies like \cite{Meng2017}, \cite{Chen2018b}, \cite{Chen2019}, \cite{Hulth2003}, \cite{Chen2018a} and \cite{Ye2018} to have the predicted and golden keywords properly compared during evaluation. A stemmer that is reported is the one of \cite{Porter2006}. There are still a few works like \cite{Zhang2018} that do not report to use stemming or any other transformation in the evaluation step.
The motivation or objective of the authors is the same in most of the studies: producing meaningful and accurate keyphrases that are similar to those set by humans which are used as ground-truth. Besides that, there are a few studies such as \cite{Misawa2019} or \cite{Chen2018a} that aim for a higher diversity or avoiding duplicates in the produced keyphrases. Producing a different number of keyphrases for each document is another requirement. It was met just recently by the model of \cite{Yuan2018}.

Overcoming the need for labeled or domain-specific data was also important for certain studies like \cite{Ye2018} and \cite{Nart2014}. Few works such as \cite{Rose2010} and \cite{Zhang2017} focus on computational efficiency and generation speed while trying to keep state-of-the-art accuracy. Other works such as \cite{Chan2019} and \cite{Villmow2018} are based on neural networks and attempt to generate more keyphrases (the former) or automate feature crafting (the latter). Finally, \cite{Boudin2016} creates a framework for implementing popular methods instead of proposing a new one.

%%% 
All studies do perform a formal evaluation of their contribution with the exception of \cite{Campos2018} where they highlight the functional features of their method by means of a practical demonstration. In the evaluation phase, they usually compare with similar methods used as baselines. 
Regarding the choice of baselines, we observed a similar trend in both extractive and abstractive KG studies. The earlier extractive works such as \cite{Witten1999}, \cite{Nguyen2010} or \cite{Bhowmik2008} do not compare against other methods. In few cases such as in \cite{Hulth2003} and \cite{Wan2008}, they compare different versions (or configuration choices) of their basic method.

The more recent extractive works like \cite{Witten1999}, \cite{Rose2010,Mihalcea2004}, \cite{Krapivin2010}, \cite{Villmow2018}, \cite{Gollapalli2014}, and \cite{Florescu2017} compare against the earlier ones. Similarly, the earlier abstractive KG studies such as \cite{Meng2017} and \cite{Zhang2016} compare against extractive methods only. 
Instead, some of the latest abstractive works such as \cite{Chen2018b}, \cite{Zhang2018} or \cite{Chan2019} compare against both extractive and abstractive KG methods.

\subsection{Keyphrasing vs. Summarizing} 
\label{ssec:kgvsts}
Some interesting research patterns we observed are related to the strict analogy between the dynamics of TS and KG research in the last two decades. Extensive research began in the late 90s on both tasks. Early TS works were mostly extractive, same as the KG works of the same time (commonly called KE studies). They were usually based on lexical resources and features, clustering algorithms and similarity measures (e.g., \cite{Barzilay1997}, \cite{Azzam1999} or \cite{Goldstein2000}).   
Several supervised TS works such as \cite{Fukumoto2004} and \cite{Wong2008} or graph-based TS works like \cite{Wan2006}, \cite{Mani1997} and \cite{Erkan2004} bloomed, in full analogy with the KG works of Sections~\ref{ssec:supervised}~and~\ref{ssec:graph}. 

The same development path has been followed in the case of abstractive studies as well. The encoder-decoder framework equipped with attention was first used by \cite{Rush2015} for title generation. In analogy with the studies of Section~\ref{ssec:enhanced}, many studies like \cite{Chopra2016} or \cite{Nallapati2016} added copying mechanism whereas \cite{See2017} was the first that used coverage. All these innovations significantly improved the results. 
The trend towards the RL approach makes no exception. It was first introduced in text summarization studies like \cite{Paulus2017} and \cite{Chen2018fast2}. As described in Section~\ref{ssec:rl}, It has been applied in AKG just recently.

There are still a few differences between TS and KG research that are related to the nature of these tasks. First, as presented in Section~\ref{ssec:scidata}, KG research works have mostly used scientific paper data. TS studies, on the other hand, have been mostly based on news articles (e.g., \cite{McKeown1995}, \cite{Kaikhah2004} or \cite{Harabagiu2005}). In fact, most of the popular TS datasets like those described in \cite{Grusky2018}, \cite{David2003}, and \cite{Nallapati2016} are made up of online news articles preprocessed by the authors.     

Another difference lies in the metrics that are used to perform the evaluation of the two tasks. KG methods are usually assessed by means of F$_1$ and recall whereas TS studies use more complex scores such as ROUGE of \cite{Lin2004} or sometimes even BLEU of \cite{Papieni2002}.
\section{Discussion}
\label{sec:discussion}
This study presents a survey of the earlier extractive KG methods and the recent cutting-edge abstractive ones that are based on the encoder-decoder framework. 
We first describe in brief some of the pivotal KE works which are supervised, unsupervised or graph-based. They were very successful and shaped the research field in the 2000s, mainly because of their speed and simplicity.
We then present the available keyphrase datasets that are popular in the literature and describe OAGKX, a huge article data collection that is released with this paper. It can be used as a data supplement for training deep learning models that require millions of samples. It might as well serve as a source for creating derivative datasets of scientific articles from more specific research disciplines. 
The shift to the recent abstractive methods was mainly pushed from the need to annotate documents with keyphrases that do not necessarily appear in the original text. The availability of the easy-to-implement encoder-decoder framework was another motive. Advanced mechanisms such as attention, copying and coverage were added one by one and improved not only the accuracy but also the diversity of the produced keyphrases.  

We further observed several similar patterns between TS and KG research. They include the transit from extractive to abstractive strategies, the use of technically advanced mechanisms (e.g., attention, copying, and coverage), and the reformulation of the methods from the reinforcement learning perspective. The latter trend is very promising and we expect to see many works in the near future exploring it in several ways for achieving different goals.

\section*{Acknowledgment}
\label{sec:ack}
This research work was supported by the project No. CZ.02.2.69/0.0/0.0/16 027/0008495 (International Mobility of Researchers at Charles University) of the Operational Programme Research, Development and Education, the project no. 19-26934X (NEUREM3) of the Czech Science Foundation and ELITR (H2020-ICT-2018-2-825460) of the EU.

\end{document}